\pdfoutput=1
\documentclass[11pt]{article}
\usepackage[hyperref]{acl2023}

\usepackage{times}
\usepackage{latexsym}

\usepackage[T1]{fontenc}

\usepackage{CJKutf8}
\newcommand{\Korean}[1]{{\begin{CJK}{UTF8}{mj}\small{#1}\end{CJK}}}
\newcommand{\Japanese}[1]{{\begin{CJK}{UTF8}{min}\small{#1}\end{CJK}}}

\usepackage[whole]{bxcjkjatype} 
\usepackage[unicode]{hyperref} 

\usepackage[utf8]{inputenc}

\usepackage{microtype}

\usepackage{graphicx}
\usepackage{multicol}
\usepackage{linguex}
\usepackage{footnote}
\makesavenoteenv{tabular}
\makesavenoteenv{table}
\usepackage[inline]{enumitem}

\usepackage{tikz}

\usepackage{amsmath}

\title{Politeness Stereotypes and Attack Vectors: Gender Stereotypes in Japanese and Korean Language Models}

\author{Victor Steinborn \and Antonis Maronikolakis  \and Hinrich Sch\"utze  \\
       Center for Information and Language Processing (CIS), LMU Munich, Germany \\
       Munich Center for Machine Learning (MCML), Germany \\
       \texttt{\{vsteinborn, antmarakis\}@cis.lmu.de} \\}

\newcounter{notecounter}
\newcommand{\enotesoff}{\long\gdef\enote##1##2{}}
\newcommand{\enoteson}{\long\gdef\enote##1##2{{
\stepcounter{notecounter}
{\large\bf \hspace{1cm}\arabic{notecounter} $<<<$ ##1: ##2 $>>>$\hspace{1cm}}}}}
\enoteson
\enotesoff

\DeclareTextCommandDefault{\textregistered}{\textcircled{%
      \check@mathfonts\fontsize\sf@size\z@\math@fontsfalse\selectfont R}}

\long\def\eat#1{}

\begin{document}
\maketitle
\begin{abstract}
        In efforts to keep up with the rapid progress and use of large language models,
        gender bias research is becoming more prevalent in NLP.
        Non-English bias research, however, is still in its infancy with most work focusing on English.
        In our work, we study how grammatical gender bias relating to politeness levels manifests in Japanese and Korean language models.
        Linguistic studies in these languages have identified a connection between gender bias and politeness levels, 
        however it is not yet known if language models reproduce these biases.
        We analyze relative prediction probabilities of the male and female grammatical genders using templates and find
        that informal polite speech is most indicative of the female grammatical gender,
        while rude and formal speech is most indicative of the male grammatical gender.
        Further, we find politeness levels to be an attack vector for allocational gender bias 
        in cyberbullying detection models. Cyberbullies can evade detection
        through simple techniques abusing politeness levels.
        We introduce an attack dataset to
        \begin{enumerate*}[label={(\roman{*})}]
        \item identify representational gender bias across politeness levels,
        \item demonstrate how gender biases can be abused
        to bypass 
        cyberbullying detection models and
        \item show that allocational biases can be mitigated via training on our proposed dataset.
        \end{enumerate*}
        Through our findings we highlight the importance of bias research moving beyond its current English-centrism.
\end{abstract}

\section{Introduction}

Gender Bias research in NLP overwhelmingly focuses on the English language 
\cite{kaneko-etal-2022-gender, bartl-etal-2020-unmasking, liang-etal-2020-monolingual, steinborn-etal-2022-information}, 
despite only 5\% of the world's population being native speakers \cite{central_intelligence_agency_world_2022}. 
With the continued development and adoption of NLP technologies,
we expect fair NLP systems in non-English languages to have greater precedence in the future.
In order to ensure that language models courteously portray individuals and fairly allocate resources 
and opportunities \cite{blodgett-etal-2020-language,blasi-etal-2022-systematic}, greater effort should be spent studying 
the rich assortment of linguistic features present in the world's languages.

In this paper, we will study one of these culturally-specific linguistic features: politeness levels in Korean and Japanese.
Linguistic studies on politeness note the presence of gender stereotypes in these languages, where women, compared to men, 
are generally expected to speak in more polite speech levels and use more honorifics (all collectively referred to as \textbf{politeness levels} in this study) 
\cite{okamoto_variability_2013, tsujimura_sociolinguistics_2017, sung-yun_womens_1983}. However, research in this area has been restricted to linguistics, 
while the consequences for NLP remain unexplored.

\eat{In light of these linguistic studies and the biases they highlight, we see this as an important problem to consider in NLP,
not only because it has been shown that non-gendered words can signal learned gender biases
 \citep{bolukbasi_man_2016, kaneko-etal-2022-gender, gonen-goldberg-2019-lipstick-pig}.}

Using these linguistics studies as a springboard,
we can avoid pitfalls arising from English-centric approaches to bias analysis. While important and pioneering work, recent research on gender bias in 
non-English languages limits its reach by only translating from English to a target language, resulting in much nuance concerning gender bias lost in translation
\citep{steinborn-etal-2022-information, kaneko-etal-2022-gender, camara-etal-2022-mapping, bartl-etal-2020-unmasking}.
For example, consider the following pair of Japanese sentences:
     \a. \Japanese{勉強する。}
     \b. \Japanese{勉強します。}

Both of these sentences can be translated to 
``(I) will study.'' (Japanese is a null-subject language, here the subject ``I'' is assumed), however \textit{sentence a} is more informal 
and tends to be used predominantly among family and friends, whereas \textit{sentence b} is more polite, and is more appropriate to use among 
acquaintances and in polite company \citep{eri_genki_2011}.
This subtlety would be missed without any linguistically-motivated analysis.

In our work, we examine if these politeness-level biases are in fact learned by language models and the consequences this phenomenon might entail.
Specifically, using the terminology introduced in \citet{blodgett-etal-2020-language}, we examine popular transformers trained with the 
Masked Language Modeling task \citep{devlin-etal-2019-bert} for representational gender biases relating to politeness levels, and consequently, 
how allocational biases manifest by studying cyberbullying detection models.
Namely, we study the prediction probabilities of the Japanese and Korean equivalents of the third person pronouns ``he'', ``she'', as well as several 
gender-neutral alternatives using large language models.

To investigate gender bias,
we propose a novel template structure to evaluate which of the politeness levels the grammatical male and female gender \textit{use}
(via a \textit{speaker}), as well as 
the politeness levels that are used to \textit{speak of} them
(via a \textit{narrator}). 
We do this by using sentences of the following form in Korean and Japanese: 
\ex. \{mask\}\ said ``(I) will \{verb\}.''

where the utterance of the speaker (\{mask\}) is directly quoted by a narrator, and politeness levels are encapsulated in the verbs (i.e., in the verbs ``to say'' and \{verb\}). 

For our representational bias study, we find statistically significant differences between politeness levels for both 
speakers and narrators, namely that the politeness level most associated with the female grammatical gender is the polite informal level, for both speakers and narrators. 
On the other hand, the male grammatical gender is comparatively most associated with formal and rude speech for the speaker, while for the narrator the male grammatical gender 
is most associated with rude speech in Japanese and the presence of honorifics (i.e., when the male grammatical gender is elevated) in Korean, propagating gender stereotypes.

Considering these biases, we investigate a popular Japanese cyberbullying detection model for allocational biases.
Namely, we prepare an attack dataset, where we append information concerning the grammatical gender of the target of the tweet to each example of a Japanese toxic tweet dataset, 
\footnote{\href{https://www.surgehq.ai/datasets/japanese-hate-speech-insults-and-toxicity-dataset}{https://www.surgehq.ai}} utilizing different politeness levels. We find a large gender discrepancy when honorific language is used:
the calculated F1 scores for the grammatical female gender are substantially lower than the grammatical male and gender-neutral forms.

Finally, we demonstrate how we can mitigate gender bias, achieve state-of-the-art performance and protect against politeness-based attacks through training a multilingual 
SentenceBERT \citep{sentencebert} transformer via few-shot learning on our proposed dataset.

We publish our code and dataset.\footnote{\href{https://github.com/VSteinborn/politeness-attacks}{https://github.com/VSteinborn/politeness-attacks}}
\textbf{In summary}, our findings are:
\begin{enumerate*}[label={(\roman{*})}]
        \item We observe a prevalent male bias across all politeness levels in language models.
        \item We determine polite informal politeness levels are most associated with the female gender,
        while formal, rude and honorific politeness levels are most associated with the male gender.
        \item We identify stereotypical associations between gender and male/female-dominated spaces.
        \item We observe substantial gender differences in Japanese cyberbullying detection efficiency when models are attacked with honorific language.
        \item We mitigate these biases by few-shot training SentenceBERT on a modified version of our attack dataset.
\end{enumerate*}

\section{Related Work}

\textbf{Gender bias} in NLP is most commonly 
studied within the context of English \citep{steinborn-etal-2022-information, kaneko-etal-2022-gender, camara-etal-2022-mapping, bartl-etal-2020-unmasking}, with other languages less commonly studied. Research in non-English settings is predominantly done in multilingual contexts, where non-English texts are treated
as translations of English source text. This approach ignores language-specific features or conventions. 
For example, \citet{bartl-etal-2020-unmasking} directly translated templates from English to German, which do not perform as well, due to the fact that the templates were not designed to respect 
German gender agreement rules, a fact \citet{steinborn-etal-2022-information} 
attempted to address by using the pair-like structure of CrowS-Pairs \citep{nangia-etal-2020-crows}.
In another study, \citet{kaneko-etal-2022-gender} translated the CrowS-Pairs dataset \citep{nangia-etal-2020-crows} to Japanese, 
however, gender information was lost about 40\% of the time.

Common topics on gender bias are occupational stereotypes \citep{caliskan_semantics_2017, arteaga_bias_2019, bartl-etal-2020-unmasking} and 
general representational stereotypes \citep{nadeem-etal-2021-stereoset, nangia-etal-2020-crows, nissim-etal-2020-fair}.

Several studies consider Korean and Japanese. 
\citet{kaneko-etal-2022-gender} proposed a multilingual technique, which uses parallel texts to probe pretrained 
models for gender bias. However Japanese-specific features of the language were not exploited in their proposed method.
\citet{cho-etal-2019-measuring,prates_assessing_2018} test commercial machine translation systems.
In a supplementary experiment, \citet{cho-etal-2019-measuring} attempted to examine if politeness affects gender 
bias, however only the informal -\Korean{해} and polite -\Korean{해요} forms 
were considered and no significant change in gender bias signals are detected.

Our study examines if there is in fact a correlation between gender bias 
and politeness levels and, to the best of our knowledge, is the first in-depth study on modern NLP models in this domain.

\textbf{Politeness} 
research is primarily focuses, almost exclusively, on English.
A popular topic in politeness research are direct requests in Wikipedia edit requests, inspired by the seminal work of \citet{danescu-niculescu-mizil-etal-2013-computational}. 
Politeness in direct requests have since been studied in predicting politeness using neural networks \citep{aubakirova-bansal-2016-interpreting}, 
and other languages, including Korean \citep{srinivasan_tydip_2022}.

With regards to politeness, gender bias is not usually the focus, 
however \citet{danescu-niculescu-mizil-etal-2013-computational} does mention female Wikipedians 
were found to be generally more polite, in agreement with prior linguistic studies \citep{herring_compcuture-1994}.

Automated detection of \textbf{hate speech and cyberbullying} has become more prevalent with the 
increased use of social media and online platforms \cite{vidgen-etal-2019-challenges}. 
While early work focused predominantly on English \cite{waseem,davidson,founta}, work to develop benchmarks, datasets and models for other languages is 
rising \citep{mishra_tackling_2019, rottger-etal-2022-multilingual,multi_hate_speech,hatespeech_crosslingual_embds,maronikolakis-etal-2022-listening,yuan-etal-2022-separating,refugee_crisis_german,hatespeech_zero_shot_cross_lingual}.

Despite progress, hate speech models and datasets are prone to certain pitfalls, 
such as low generalization abilities, biased data and inconsistent 
definitions \cite{hatecheck,hate_speech_data_analysis,hate_speech_cross_dataset,hatespeech_biased_datasets}. 
Further, vulnerabilities of hate speech models against adversarial attacks have been uncovered.
\citet{grondahl_all_2018} demonstrated how appending the word ``love'' rendered tested models ineffective. In our work, we investigate vulnerabilities against politeness-level attacks.

This raises the question whether a user peddling hate speech online could use language-specific 
biases to evade detection, or conversely, whether a designer of detection systems could 
leverage this knowledge to enhance the model's performance. In our work, 
we attempt to answer these questions for cyberbullying identification in Japanese.
Namely, we analyze models for biases relating to politeness levels and propose a linguistics-oriented 
solution to better prepare models against adversarial attacks.

\textbf{Few-shot learning} is emerging as a popular trend in the NLP community, 
built on the emergent abilities of large pretrained language models \cite{wei2022emergent}, 
which have been shown to work well in few- and zero-shot settings 
\cite{gpt3,gao-etal-2021-making,sanh2022multitask,le-scao-rush-2021-many,https://doi.org/10.48550/arxiv.2204.14264}. 
Few-shot learning has benefited from the use of prompting \citep{schick-schutze-2021-exploiting}, which has 
been shown to be competitive with models orders of magnitude larger \cite{schick-schutze-2021-just}. While prompting is a useful technique to aid in model learning, it requires manual crafting of prompts and labels. 
While there has been work to improve prompting, it remains a noisy 
process \cite{schick-etal-2020-automatically,logan-iv-etal-2022-cutting,lu-etal-2022-fantastically,shin-etal-2020-autoprompt,zhao-schutze-2021-discrete,https://doi.org/10.48550/arxiv.2205.11822,10.1145/3491102.3517582,mishra-etal-2022-reframing}.

Recently, SetFit \cite{setfit} introduced a prompt-free approach to few-shot learning. Through the use of 
SentenceBERT and its Siamese-network training paradigm \cite{sentencebert}, SetFit 
generates pairs of training examples and learns to minimize the distance of representations of training examples 
of the same class and, conversely, to maximize the distance for examples from different classes. This process results in 
a model that can generate strong sentence embeddings, which can be then used to train a classification head on a task.

\section{Methodology}

\begin{table*}[h]
        \centering
        \begin{tabular}{l l l}
                \hline
                Lang. & Template & Application\\
                \hline
                Ja & \{mask\}\Japanese{は「}\{speakerNoun\}\{speakerVerbEnding\}\Japanese{」と}\{narratorVerb\} \Japanese{。} & Rep. \\
                Ko & \{mask\}\Korean{은}/\Korean{는} ``\{speakerNoun\}\{speakerVerbEnding\}''(\Korean{이})\Korean{라고} \{narratorVerb\}. & Rep.\\
                Ja & \{tweet\}\Japanese{(}\{genderTerm\}\Japanese{はこう}\{narratorVerb\}\Japanese{)} & Allo. (train)\\
                Ja & \{tweet\}\Japanese{(}\{genderTerm\}\Japanese{にこう}\{narratorVerb\}\Japanese{)} & Allo. (test)\\
                \hline
        \end{tabular}
        \caption{Templates used to probe representational (Rep.) and allocational (Allo.) biases.}
        \label{tbl:rep-bias-temp}
\end{table*}

\subsection{Representational Biases}

We probe Masked Language Models (MLMs) for representational biases (i.e., biases relating 
to how different persons are portrayed by NLP models \citep{blodgett-etal-2020-language}), 
using a novel template approach. The templates, shown in Table \ref{tbl:rep-bias-temp}, are designed such that we can probe both
the type of language (rude, polite, formal and honorific) 
the models associate with different individuals (via a speaker) and the language that is used to speak of these individuals (via a narrator).

\textbf{Templates.}
To simplify presentation of experiments,
we differentiate between 
a so-called \textit{speaker} and a so-called \textit{narrator}.
The \textbf{speaker} will speak of an action using a \Japanese{する} (Japanese) or a \Korean{하다} verb (Korean).
\Japanese{する} and \Korean{하다} verbs consist of a noun (\{speakerNoun\}) 
and the verb ``to do'' (\{speakerVerbEnding\}) where the verb can change with the politeness level \citep{roh_korean_2013, eri_genki_2011}.

These pieces can simply be prepended with the noun (\{speakerNoun\}\{speakerVerbEnding\}). 
For example, the verb for ``to study'', may be formed via the noun \Japanese{勉強} in Japanese and 
\Korean{공부} in Korean (\{speakerNoun\}), and combined with the informal form of the verb ``to do'', namely \Japanese{する} in Japanese and 
\Korean{해} in Korean (\{speakerVerbEnding\}), to form \Japanese{勉強する} and \Korean{공부해} respectively \citep{roh_korean_2013, eri_genki_2011}.

What makes \Japanese{する} and \Korean{하다} verbs particularly appealing for templates is that politeness is 
encapsulated in the verb (\{speakerVerbEnding\}) and that the noun (\{speakerNoun\}) can be freely exchanged between 
politeness levels, a fact also exploited by \citet{cho-etal-2019-measuring} when working with Korean.

To complete the template, the \textbf{narrator} directly quotes the utterance of the speaker, $X$, via the Japanese and Korean 
equivalent of ``\{mask\} said `$X$'.''. We then let the model predict the gender identity of the speaker via a mask token (\{mask\}).
The templates are shown in Table \ref{tbl:rep-bias-temp}, where \{narratorVerb\} is the verb 
``to say'' in the past tense form at various politeness levels.

\textbf{Data.} \Japanese{する} and \Korean{하다} verbs are taken from standardized language proficiency tests.
We used 142 \Japanese{する} verbs from the JLPT \footnote{\url{https://www.jlpt.jp/e/}} and 107 \Korean{하다} 
verbs from the TOPIK \footnote{\url{https://www.topik.go.kr}}. The reasoning behind using verbs from language proficiency tests
is that they are common and standardized.

We convert the verbs into common politeness levels used in each language, as outlined in Table \ref{tbl:polite-levels}. 
Politeness levels are used to indicate differing levels of politeness, formality or respect 
towards a subject (via honorifics) or listener \citep{roh_korean_2013, eri_genki_2011, keigo_note_2014}.

\begin{table}
        \centering  
        \begin{tabular}{l l l l l}
                \hline
                Politeness Level & Ex. & P & F & H  \\
                \hline
                \eat{\small{\Japanese{〜ぞ}}}  rude\_zo   & \small{\Japanese{するぞ}} & & & \\
                \eat{\small{\Japanese{〜ぜ}}}  rude\_ze   & \small{\Japanese{するぜ}} & & & \\
                \eat{\small{\Japanese{基本形}}} plain    & \small{\Japanese{する}}   & & & \\
                \eat{\small{\Japanese{丁寧語}}} teineigo & \small{\Japanese{します}} & $\star$ & & \\
                \eat{\small{\Japanese{謙譲語}}} kenjōgo  & \small{\Japanese{いたす}} & $\star$ & $\star$ & \\
                \eat{\small{\Japanese{尊敬語}}} sonkeigo & \small{\Japanese{なさる}} & $\star$ & $\star$ & $\star$ \\
                \eat{\small{\Korean{해체}}} heche  & \small{\Korean{해}} & & & \\
                \eat{\small{\Korean{해요체}}} heyoche& \small{\Korean{해요}} & $\star$ & & \\
                \eat{\small{\Korean{하십시오체}}} hapsyoche  & \small{\Korean{합니다}}   & $\star$ & $\star$ & \\
                \eat{\small{\Korean{해체+시}}} heche+hon. & \small{\Korean{하셔}} &  & & $\star$ \\
                \eat{\small{\Korean{해요체+시}}} heyoche+hon.  & \small{\Korean{하셔요}} & $\star$ &  & $\star$ \\
                \eat{\small{\Korean{하십시오체+시}}} hapsyoche+hon. & \small{\Korean{하십니다}} & $\star$ & $\star$ & $\star$ \\
                \hline
        \end{tabular}
        \caption{Overview of Japanese and Korean politeness levels. 
        The verb ``to do'' (\Japanese{する} and \Korean{하다}) is used to illustrate (Ex.)
        how verbs change. 
        The general politeness (P), 
        the general formality (F) and
        the elevation of the subject performing the action via honorific language (H) is indicated across levels.
        Note: the informal rude\_ze and rude\_zo forms indicate rough speech.}
        \label{tbl:polite-levels}
\end{table}

By taking all combinations of speaker nouns, speaker verb endings and narrator verb endings, 
we have $3852 = 107 \times 6 \times 6$ sentences for Korean 
and $4260= 142 \times 6 \times (6-1)$ sentences for Japanese for the representational bias study. Note, 
the minus one in the calculation for Japanese is 
because kenjōgo can only be used to speak humbly of one's own actions, 
and thus cannot be used by the narrator \citep{keigo_note_2014}.

\textbf{Models.} All models are listed in table \ref{tbl:apx_locations} in the appendix .
We selected the ten most downloaded MLMs on Huggingface \citep{wolf-etal-2020-transformers} for each language
that have a single token for ``he'' and ``she'' each.

\textbf{Search tokens.} We search for the following tokens that could appear under the \{mask\} token, namely the terms 
``he'' (ja: \Japanese{彼}, ko: \Korean{그}), ``she'' (ja: \Japanese{彼女}, ko: \Korean{그녀}) and several demonstrative gender-neutral third-person pronouns. 

For Japanese, we search for the gender-neutral formal proximal, medial and distal pronouns ``\Japanese{こちら}'', 
``\Japanese{そちら}'' and ``\Japanese{あちら}'' respectively, as well as their informal versions ``\Japanese{こいつ}'', 
``\Japanese{そいつ}'' and ``\Japanese{あいつ}''. For Korean, we follow \citet{cho-etal-2019-measuring} and search for 
``\Korean{걔}'' and ``\Korean{그 사람}''. 

\textbf{Correlations of stereotypically mono-gender dominated locations} were also investigated.
To our understanding, this is the first study that investigates location biases in large language models.

To investigate correlations between gender and locations,
we prepend our representational bias templates with (\{location\}\Japanese{で}) in Japanese and (\{location\}\Korean{에서}) 
in Korean, which translates to ``(at \{location\})'', which we use to give context on the location of the scene \citep{roh_korean_2013, eri_genki_2011}.

We chose ten locations for the male and female grammatical gender based on
surveys about gender inequality in Japan and South Korea 
\citep{world_economic_forum_global_2021, gender_equality_bureau_cabinet_office_current_2022, korean_womens_development_institute_2021_2022, korean_womens_development_institute_is_digital_2022} 
and discussions with native speakers, who corroborate our choices with their lived experiences.
The full list of locations can be found in table \ref{tbl:apx_locations} in the appendix. 
Male locations are generally associated with positions of authority and manual labour, whereas female locations 
are associated with health and childcare.

\subsection{Allocational Biases}
We test for gender differences in allocational biases (i.e., biases relating to how resources are 
allocated \citep{blodgett-etal-2020-language}), by investigating toxic content detection differences when models are attacked via politeness-level manipulations.

Namely, we compare performance of the most downloaded\footnote{With over 500 downloads per month on Huggingface \citep{wolf-etal-2020-transformers} at the time of writing.} Japanese cyberbullying detection model \citep{shibata2022yacis-electra} against our proposed model, which is designed 
to jointly detect cyberbullying and protect against politeness-level attacks.

The baseline model was pretrained on the YACIS corpus \citep{ptaszynski2012yacis} and fine-tuned on the Harmful BBS 
Japanese Comments Dataset \citep{ptaszynski2018automatic, ja-cyberbullying-dataset} and the Twitter 
Japanese Cyberbullying Dataset \citep{ptaszynski2012yacis}.

We use a different, recently released, balanced (50/50 split) toxic tweet dataset from 
Surge AI,\footnote{\href{https://www.surgehq.ai/datasets/japanese-hate-speech-insults-and-toxicity-dataset}{https://www.surgehq.ai} (Dataset created: 2022.07.02)}
a professional data labeling platform, to test models for allocational biases.

We fed the tweets into the \{tweet\} slot in the templates under 
the ``Allo.'' application column in Table \ref{tbl:rep-bias-temp} (where the search terms 
from earlier are substituted in the \{genderTerm\} slot). 
We test both models on the test template in Table \ref{tbl:rep-bias-temp} 
(translation: ``\{tweet\} (it was told so to \{genderTerm\})''), which serves to give information of the victim of the potentially toxic tweet. 
All possible combinations of tweets, gender terms and politeness levels constitute our attack dataset, 
which consists of $39,160$ sentences ($= (987-8)$ unique tweets $\times 8$ gender terms $\times (6-1)$ politeness levels (via \{narratorVerb\}))
in total. Note, eight tweets are used for our few-shot learning setup and kenjōgo was removed since it cannot be used by the narrator (similarly to our representational bias experiments).

Finally, we use the training template in 
Table \ref{tbl:rep-bias-temp}  (translation: ``\{tweet\} (\{genderTerm\} said it so)'') to
train our model, using few-shot learning. For further experimental details, refer to appendix \ref{sec:exp_setup}.

\begin{figure*}
        \includegraphics[width=\linewidth]{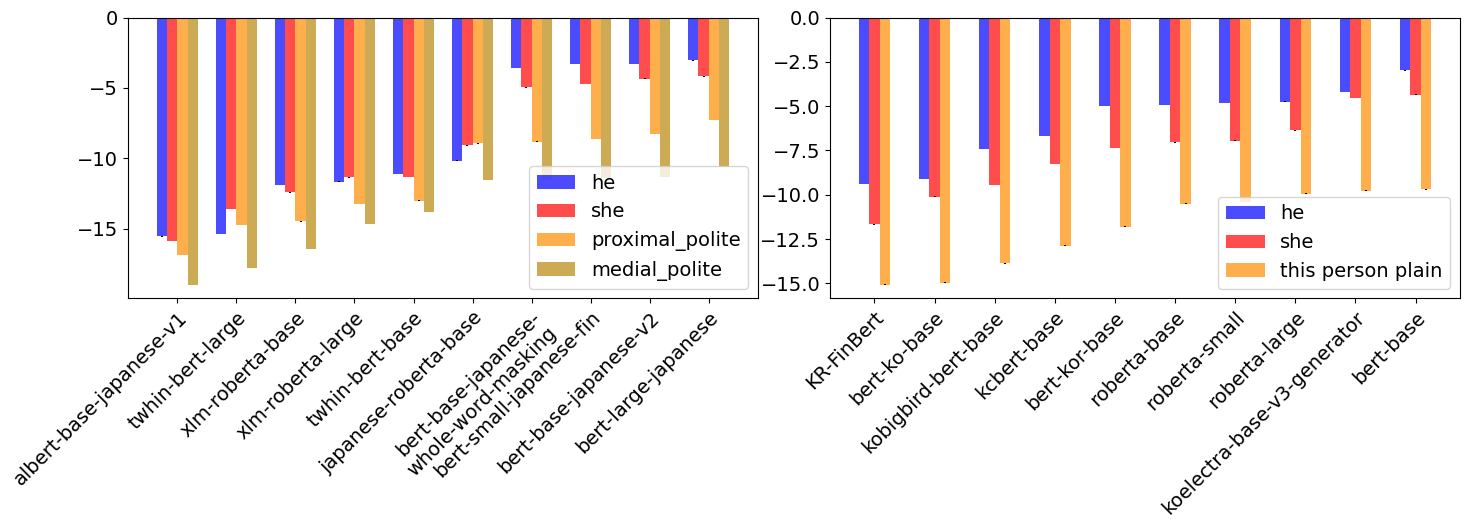}\par 
        \caption{Japanese (left) and Korean (right) log probabilities of the search tokens for each model. 
        More negative log probabilities correspond to lower prediction probabilities. ``he'' is more likely than 
        ``she'' in all Korean models and seven out of ten Japanese models. The gender-neutral tokens are the least likely in all Korean 
        models and in eight out of ten Japanese models. Standard errors are shown (their magnitudes are at most 1\% of the mean).}
        \label{fig:gender-tokens}
\end{figure*}

\subsection{Few-Shot Learning}
For our proposed method, we are introducing a modified dataset that aids in training the model against politeness-level attacks to evade cyberbullying detection in Japanese.
We use the SetFit \cite{setfit} framework to train a multilingual SentenceBERT model\footnote{\url{https://huggingface.co/sentence-transformers/paraphrase-multilingual-mpnet-base-v2}} that was pretrained on (among other languages) Japanese data.

Out of the total 987 original tweets, we used 8 tweets plus 384 template-modified examples of these tweets for a total of 392 training examples. The 8 tweets selected for training were removed from the original dataset and the template used for training was not used for the generation of politeness-attack tweets. Thus, we ensure no overlap between training and evaluation data.

With SetFit, the model is trained in a contrastive learning manner: given two 
training examples, the model learns to decrease representation distance (e.g., cosine similarity) between 
them if they belong to the same class and increase distance between them if they belong to  different classes.

\section{Results and Analysis}

\subsection{Representational Bias}

In our representational bias experiments, we find that
\textbf{``he'' is the most likely form of address, while gender-neutral pronouns are the least likely.}
We observe this effect by comparing the distribution of log probabilities of the tokens 
``he'', ``she'' and the gender-neutral search tokens under the mask 
(distributions of gender tokens shown in Figure \ref{fig:gender-tokens}). Apart from the male and female pronouns, the only gender-neutral search tokens with high probability are the polite proximal and 
polite medial demonstrative pronouns (\Japanese{こちら} and \Japanese{そちら}, respectively) for Japanese, 
and the casual demonstrative pronoun \Korean{걔} for Korean.
Generally, we observe the ``he'' token has a higher probability than the ``she'' token, with gender-neutral tokens being even less likely.

\begin{figure}
        \includegraphics[width=\linewidth]{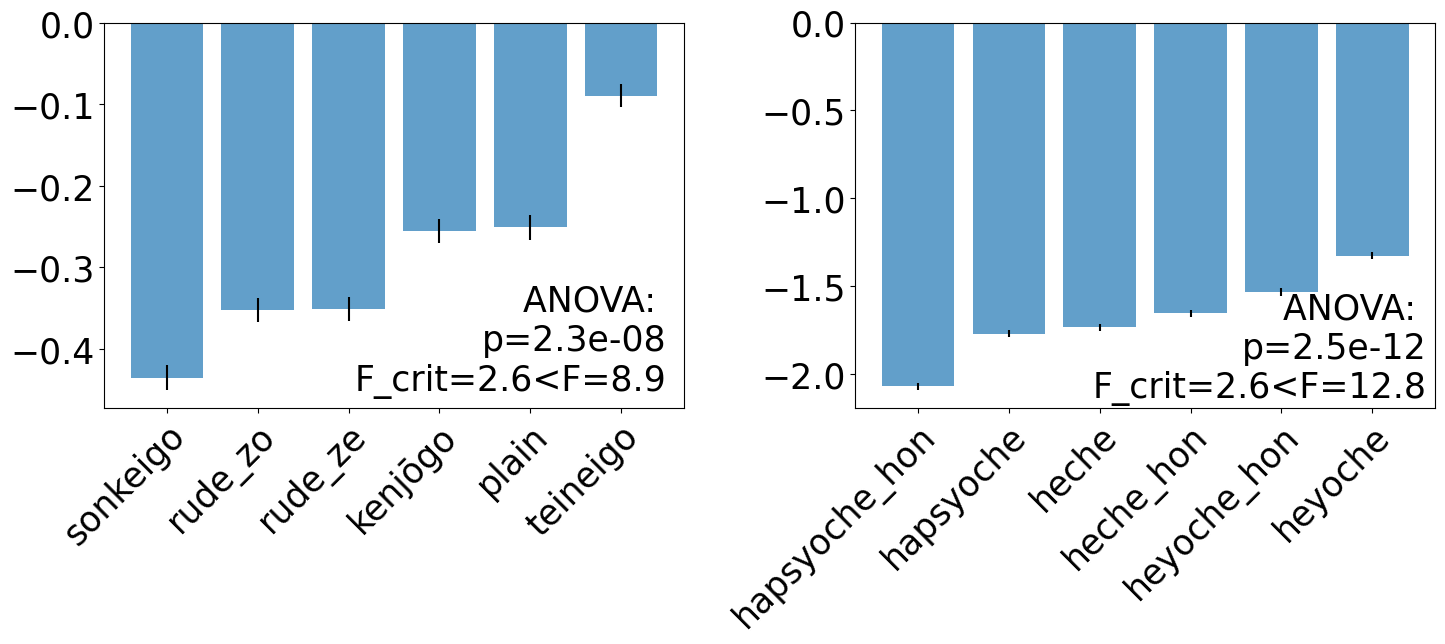}\par 
        \caption{Japanese (left) and Korean (right) mean log probability differences between ``she'' and ``he'' across speaker levels.
        Negative scores indicate more male-biased predictions. Standard errors are shown.
        }
        \label{fig:speaker-levels}
\end{figure}

Further, we identify that \textbf{female speakers most likely speak in an informal polite level, while male speakers are more rude or formal.}
We observe representational biases within the model by taking the average of the difference between the logs of the prediction probabilities of ``she'' 
and ``he'' under the mask (mathematically, $\log{p(\text{{mask}=she})}-\log{p(\text{{mask}=he})}$), 
across all sentences. Figure \ref{fig:speaker-levels} presents the results.

We first verify that the differences of log probabilities across speaker levels are (roughly) normally distributed and the variances of log probabilities across speaker levels are of similar sizes.
Then, we perform 
ANOVA (Analysis of Variation, \citet{snedecor_statistical_1996}) and reject (via a statistical F- and p-test) the null hypothesis of all averages between politeness levels 
being equal with $p=2\times 10^{-8}$ and $F_{crit.} = 2.6 < F=8.9$ for Japanese and $p=3\times 10^{-12}$ and $F_{crit.} = 2.6 < F=13$ for Korean, assuming 
a significance level $\alpha=0.05$.

We observe the largest differences between sonkeigo (honorific speech; most male-biased) and teineigo 
(informal polite speech; most female-biased) in Japanese and hapsyoche (formal language; most male-biased) with an honorific marker and heyoche 
(informal polite speech; most female-biased) in Korean. Additionally, we observe negative averages across all politeness levels, 
indicative of a general male bias within the models. Thus, we conclude that biases associating female speech with 
informal polite speech, and male speech with both formal and rude speech do exist within the studied language models.

\begin{figure*}
        \includegraphics[width=\linewidth]{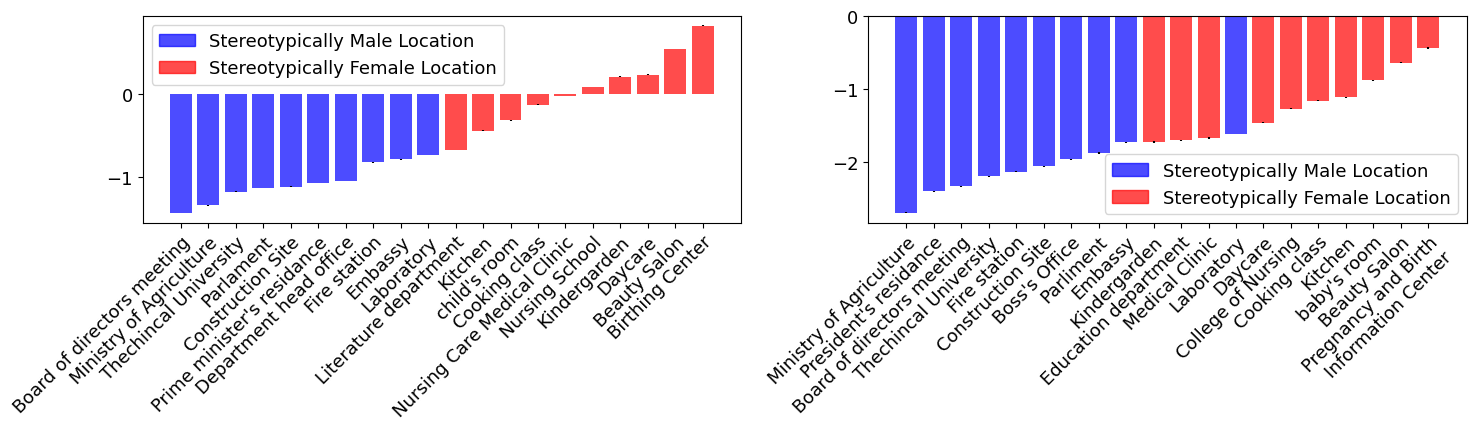}\par
        \caption{Japanese (left) and Korean (right) gender associations with locations. The vertical axis shows the mean difference between 
        female and male token prediction probabilities for each of the locations. Models assign 
        more negative score (i.e., a higher probability of predicting the ``he'' token over the ``she'' token) for stereotypically male-dominated spaces (in blue), while assigning more positive scores to female dominated spaces (in red).  
        Standard errors are shown (their magnitudes are roughly 1\% of 
        the mean).}
        \label{fig:locations}
\end{figure*}

\begin{figure}
        \includegraphics[width=\linewidth]{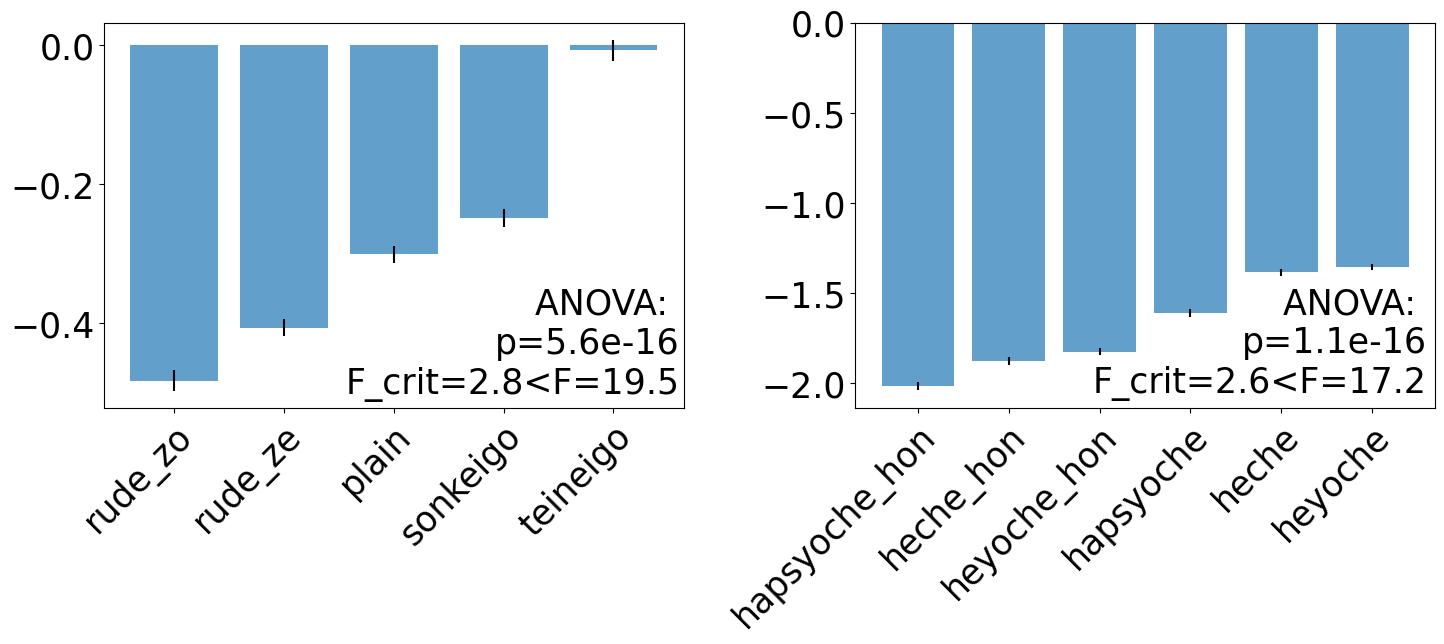}\par 
        \caption{Japanese (left) and Korean (right) mean log probability differences between ``she'' and ``he'' across narrator levels.
        Negative scores indicate more male-biased predictions. Standard errors are shown. 
        }
        \label{fig:narrator-levels}
\end{figure}

We also demonstrate that \textbf{narrators speaking of the female gender tend to use informal polite levels, while honorific and rude language is used for the male gender.}
Similarly to analyzing speaker levels, we examine variations between \textit{narrator levels} via 
differences in the logs of the prediction probabilities of the tokens for ``she'' and ``he''. Results 
are shown in Figure \ref{fig:narrator-levels}. The kenjōgo politeness level in 
Japanese can only be used to speak humbly of one's own actions, and is thus omitted in this analysis.

After verifying we have normal distributions for each speaker level with variances of similar magnitudes, we perform ANOVA and reject the null hypothesis
that all averages between politeness levels are equal with $p=6 \times 10^{-16}$ and $F_{crit}=2.8 < F=20$ for Japanese, 
and $p=1 \times 10^{-16}$ $F_{crit}=2.6 < F=17$ for Korean, at a significance level $\alpha=0.05$.

We observe 
the largest distance between the rude\_ze and rude\_zo forms (rough speech; most male-biased speech) and 
teineigo (informal polite speech; most female-biased) for Japanese, and for Korean we see the largest 
difference between heyoche (informal polite speech; most female-biased) and 
hapsyoche (formal language; most male-biased) with an honorific marker. We note that for Korean,
the largest predictor of pro-male bias is the use of honorifics. In other words when a person is the subject of respect and social distinction, 
the model is most likely to predict the male grammatical gender. We do not see this effect in the Japanese results of this experiment.

We conclude that the female grammatical gender is more likely to be spoken of in a polite and informal tone, whereas the male gender is spoken of in tones either rude or formal.

Using the modified location templates, \textbf{we observe stereotypical associations between gender and locations}. We take the mean difference between 
the log prediction probabilities between the tokens for ``she'' and ``he'', similarly to our previous studies, and plot the differences 
across locations in Figure \ref{fig:locations}.

We note that the male-dominated spaces are male-biased (heavily negative scores).
The female dominated spaces,
while they are less male-biased than male locations, still exhibit predominantly negative scores. In Korean especially, all female locations have negative scores. 
This effect is less pronounced in Japanese with half of female locations exhibiting female-bias.

Thus, we conclude that gender bias associated with 
stereotypically mono-gender dominated spaces is present in language models, however, we note that their effect may be dwarfed by the general leverage 
of male bias.

\subsection{Allocational Bias}

\begin{table*}
        \centering
        \small
        \begin{tabular}{lcccccc|c}
                \hline
                {} &  rude\_zo &  rude\_ze &  plain &  teineigo &  kenjōgo &  sonkeigo & gender\_only \\
                \hline
                he               &       .20 $\to$ .87&       .20 $\to$ .86&          .20 $\to$ .83&             .20 $\to$ .82&            .20 $\to$ .83&             .99 $\to$ .83  & .69 $\to$ .82\\
                she              &       .20 $\to$ .87&       .20 $\to$ .86&          .20 $\to$ .83&             .20 $\to$ .82&            .20 $\to$ .83&             .20 $\to$ .83  & .34 $\to$ .83\\
                proximal\_p      &       .20 $\to$ .84&       .20 $\to$ .83&          .20 $\to$ .81&             .20 $\to$ .80&            .20 $\to$ .81&             .95 $\to$ .81  & .32 $\to$ .81\\
                medial\_p        &       .20 $\to$ .85&       .20 $\to$ .84&          .20 $\to$ .82&             .20 $\to$ .81&            .20 $\to$ .82&             .81 $\to$ .82  & .29 $\to$ .81\\
                distal\_p        &       .20 $\to$ .87&       .20 $\to$ .87&          .20 $\to$ .83&             .20 $\to$ .82&            .20 $\to$ .82&             .93 $\to$ .83  & .54 $\to$ .81\\
                proximal\_r      &       .20 $\to$ .88&       .20 $\to$ .87&          .20 $\to$ .84&             .20 $\to$ .82&            .20 $\to$ .84&             1.00 $\to$ .84 & .69 $\to$ .82\\
                medial\_r        &       .20 $\to$ .87&       .20 $\to$ .86&          .20 $\to$ .83&             .20 $\to$ .82&            .20 $\to$ .82&             .99 $\to$ .83  & .69 $\to$ .82\\
                distal\_r        &       .20 $\to$ .88&       .20 $\to$ .87&          .20 $\to$ .84&             .20 $\to$ .82&            .20 $\to$ .83&             1.00 $\to$ .84 & .67 $\to$ .82\\
                \hline
        \end{tabular}
        \caption{F1 scores for the baseline model (left of the arrows), and our SentenceBERT model 
        (right of the arrows) when evaluated on our attack dataset (left of the vertical line) and on the 
        gender\_only dataset (right of the vertical line). 
        For the gender\_only test,
        the gender-neutral polite terms (labeled with \_p) and rude terms (labeled with \_r) are also included. Our proposed model retains 
        comparatively high, gender-equal performance.}
        \label{tbl:f1-bias}
\end{table*}

For our allocational bias experiments, we show that \textbf{gender biases relating to honorifics may be used as an attack vector against cyberbullying models,}
thus demonstrating downstream allocational biases can lead to gender biases relating to politeness levels.

As a base test (\textit{tweet\_only}), we evaluate only on the original tweets in the test set (i.e., without modifying the tweets as per our attack approach). The baseline model \citep{shibata2022yacis-electra} has an F1 score of 0.40 
while our proposed SentenceBERT model has an F1 score of 0.82.
This serves as an initial gauge of how our examined models fare on normal cyberbullying tweets found online.
After our attack, we are expecting to see a drop of performance for the baseline model, while we are aiming for a minimal drop (or, ideally, no drop at all) for our proposed model.

Table \ref{tbl:f1-bias} shows the F1 scores when testing on our attack dataset across the different gender terms and politeness levels for the baseline and our proposed model.\footnote{We further experimented with a simplified few-shot learning model, where we only train using the original tweets (and not the data generated through our template scheme). Model performance was low and was thus omitted for brevity. We conjecture that since we only had 8 tweets at our disposal for training, low performance was expected.}
As hypothesised, the baseline model performs worse under our attack, across most politeness levels. The sole exception is sonkeigo, where there is a large gap between ``he'' and ``she'' attacks, indicating a strong gender bias. On the other hand, our proposed SentenceBERT model is robust against politeness attacks, scoring equivalently to the tweets\_only test (i.e., there is little difference before and after the attack).

As another base test (\textit{gender\_only}) we evaluate on the original tweets with only ``(\{genderTerm\})'' appended at the end (no politeness levels).
We observe (on the right side of the vertical line in Table \ref{tbl:f1-bias}) that, with the baseline model, ``he'' scores significantly higher than ``she''.
Additionally, we also note that compared to rude pronouns, polite gender-neutral pronouns generally have lower F1 scores, 
presumably because rude pronouns are more common in hate speech. For SentenceBERT we note higher performance and substantially fairer results across genders and politeness levels.

We interpret these results as a clear case where gender bias and biases relating to women generally not being the subject of honorific language (compared to men), 
manifest themselves as an allocational bias.
An attacker, as we show, can abuse this deficiency in models to evade detection and push hate speech onto an online platform.
Additionally, we argue the online presence of the attack vectors themselves further serve to propagate the harmful stereotype that women are not associated with honorific language.

\section{Conclusion}
In our work we investigate the manifestation of gender bias relating to politeness levels in language models, using a template-based setup to probe large pretrained language models.

We demonstrate (via the \textit{speaker}) that polite speech is most associated with the female grammatical gender, while formal and rude speech 
is most associated with the male gender. Additionally, we observe (via the \textit{narrator}) that the female gender was most likely to be spoken of 
using a polite informal tone, while the male gender was most likely to be spoken of using formal and honorific language (for Korean) or 
rude language (for Japanese).

Further, we observe that gender biases relating to politeness levels can also manifest in popular cyberbullying detection models, leading 
to allocational biases. We propose a method to mitigate these biases through few-shot learning on a linguistically-informed dataset, 
increasing performance and providing robustness against politeness-level and gender-based attacks.

We hope our study inspires further investigation of gender bias manifestation through linguistic features across more under-explored languages.

\section{Limitations and Ethical Considerations}
\subsection{Limitations}
In this preliminary study on the influence of politeness levels on gender bias in language models, 
we limited ourselves to a select set of verbs and basic politeness levels in Korean and Japanese.
There are, however, other classes of verbs we did not consider and there are more complex and nuanced ways of expressing 
politeness, respect and humility than the politeness levels we presented here \citep{keigo_note_2014}.

Additionally, there are other methods of demonstrating respect within these languages that does not involve 
a straightforward modification of a verb. Politeness may also be demonstrated through the choice of pronouns, 
as we have seen, but also through the use of titles and the choice of nouns (for example, 
the word for ``home'' could be ``\Japanese{家}'' or ``\Japanese{お宅}'' in Japanese and 
``\Korean{집}'' or ``\Korean{댁}'' in Korean, in casual and polite contexts respectively). Thus, 
the topic of politeness levels and its connection to gender bias is far more vast and complex than what is 
presented in this study.

\subsection{Ethical Considerations}
In this work we demonstrated representational and allocational gender biases with respect to politeness levels in NLP models. 
The release of this knowledge could potentially be exploited in practice to bypass cyberbullying detection 
systems, however, we see the release of this knowledge to be an important first step to making other NLP practitioners aware 
of this problem and how this could potentially affect their NLP systems.

Additionally, in this study we did not simply point out issues with the learned biases of modern NLP systems, but also attempted to mitigate them 
via our proposed linguistically-informed method. With the release of our dataset and code, we hope to assist NLP practitioners making their 
systems safer and more robust against attacks abusing politeness levels and gender biases, as well as to inspire future work in this area.

\section{Acknowledgements}
This work was funded by the European Research Council (grant \#740516).

\bibliographystyle{acl_natbib}
\bibliography{anthology-extract,custom}

\appendix

\section{Experimental Setup}
\label{sec:exp_setup}

For probing representational biases, all possible sentence combinations (3852 and 4260 combinations for Korean and Japanese, respectively) are fed into the selected models.
Evaluation took roughly 15 minutes using a single NVIDIA GeForce GTX 1080Ti GPU with a batch size of 64, for each language.

For the allocational biases, we use the selected cyberbullying model instead, and evaluation took roughly 5 minutes with a batch 
size of 64 on the same GPU.

For few-shot learning, the default SetFit \citep{setfit} parameters were used for epochs (set to 1) and number of sentence pairs (i.e., how many pairs to generate from one sentence; set by default to 20). A batch size of 32 was used. Training took place on the same GPU as the probing experiments (i.e., NVIDIA GeForce GTX 1080Ti). Training time is approximately 5 minutes for the entire training set.

\section{Correlation between Pro-Male Bias and Model Size}

\begin{figure}[h]
        \includegraphics[width=\linewidth]{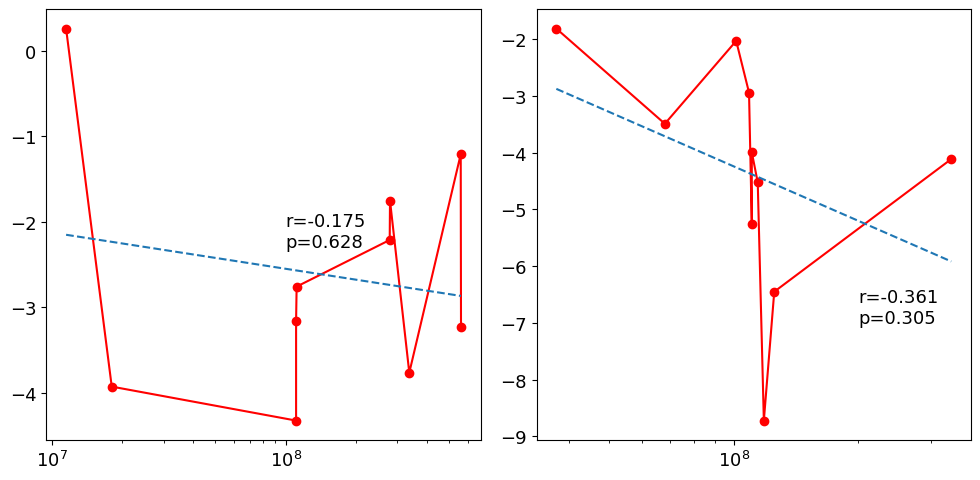}\par
        \caption{Japanese (left) and Korean (right) bias scores $s_{\text{b}}$ with parameter size. Negative scores correspond to more male-biased predictions.}
        \label{fig:parameter-size}
\end{figure}

\textbf{The correlation of gender bias with parameter count} is also investigated. 
We follow \citet{srivastava_beyond_2022} and modify the proposed social bias measure 
to our sentence templates. Namely, we calculate the bias 
score $s_{\text{b}}$, defined in equation \ref{eqn:bias}, by identifying the context $C$ (which includes the 
speaker verb and the narrator and speaker politeness levels), which minimizes the difference between the log probabilities of 
``she'' and ``he'' for each used model.

\begin{equation} \label{eqn:bias}
        s_{\text{b}} = \min_{C} \log{p(\text{mask=she}| C}) - \log{p(\text{mask=he}| C})
\end{equation}

We observe \textbf{a general correlation between trainable parameter count and pro-male-biased, with larger models exhibiting higher male bias.} We calculate $s_{\text{b}}$ via equation \ref{eqn:bias} 
and plot the variation of $s_{\text{b}}$ with the parameter count in Figure \ref{fig:parameter-size}. We observe a general trend that models become more male-biased with 
increasing parameter count, inline with the results of \citet{srivastava_beyond_2022}, however we also note that the observed correlations are not statistically significant. 
The null hypothesis, which we take to be the slope being zero, cannot be rejected with significance $\alpha=0.05$. Namely, we find $p=0.63$ for Japanese and $p=0.31$ for Korean
via\footnote{Using SciPy's \textsc{linregress} function \citep{virtanen_scipy_2020}.}
a Wald test using a t-distribution of the Wald test statistic \citep{cameron_microeconometrics_2005}.

Thus, these results are interpreted as a general trend but not as a hard rule. We expect this correlation to be 
more pronounced if we probe models with an order of magnitude larger parameter count. However, we note \citet{srivastava_beyond_2022} also observed $s_{\text{b}}$ is not monotonically 
decreasing with parameter count, thus the presence of
plateauing regions, with little correlation, cannot be ruled out.

\section{Locations and Models}

\begin{onecolumn}

\begin{table*}
\centering  
\begin{tabular}{l l l l}
        \hline
        Location & Ja & Ko & G \\
        \hline
        Parliament & \Japanese{議会} & \Korean{의회} & M \\
        \begin{tabular}{l@{}l@{}} Department head office (ja) \\ Boss's Office (ko) \end{tabular} & \Japanese{部局長室} & \Korean{사장실} & M \\
        Construction Site & \Japanese{工事現場} & \Korean{공사장} & M \\
        \begin{tabular}{l@{}l@{}} Prime Minister's Residence (ja) \\ President's Residence (ko) \end{tabular} & \Japanese{首相官邸} & \Korean{청와대} & M \\ 
        Technical University & \Japanese{工業大学} & \Korean{공과대학교} & M \\
        Board of Directors Meeting & \Japanese{取締役会} & \Korean{이사회} & M \\
        Ministry of Agriculture & \Japanese{農務省} & \Korean{농림부} & M \\
        Embassy & \Japanese{大使館} & \Korean{대사관} & M \\
        Laboratory & \Japanese{研究室} & \Korean{실험실} & M \\
        Fire Station & \Japanese{消防署} & \Korean{소방서} & M \\
        Daycare & \Japanese{保育園} & \Korean{어린이집} & F \\
        Kindergarten & \Japanese{幼稚園} & \Korean{유치원} & F \\
        \begin{tabular}{l@{}l@{}} Nursing School (ja) \\ College of Nursing (ko) \end{tabular} & \Japanese{看護学校} & \Korean{간호대학} & F \\ 
        \begin{tabular}{l@{}l@{}} Child's Room (ja) \\ Baby's Room (ko) \end{tabular} & \Japanese{子供部屋} & \Korean{아기방} & F \\ 
        \begin{tabular}{l@{}l@{}} Literature Department (ja) \\ Education Department (ko) \end{tabular} & \Japanese{文学部} & \Korean{교육학과 건물} & F \\ 
        Cooking Class & \Japanese{料理教室} & \Korean{요리교실} & F \\ 
        Kitchen & \Japanese{キッチン} & \Korean{부엌} & F \\
        Beauty Salon & \Japanese{エステサロン} & \Korean{미용실} & F \\
        \begin{tabular}{l@{}l@{}} Birthing Center (ja) \\ Pregnancy and Birth Information Center (ko) \end{tabular} & \Japanese{出産センター} & \Korean{임신출산 정보센터} & F \\ 
        \begin{tabular}{l@{}l@{}} Nursing Care Medical Clinic (ja) \\ Medical Clinic (ko) \end{tabular} & \Japanese{介護医療院} & \Korean{진료소} & F \\
        \hline
\end{tabular}
\caption{Locations used in this study. The gold-labeled stereotypical gender association (G) is indicated and is either male (M) or female (F).}
\label{tbl:apx_locations}
\end{table*}

\begin{table*}
\centering  
\begin{tabular}{l l l l}
        \hline
        Huggingface Model Name & Lang. & Params. & App. \\
        \hline
        \href{https://huggingface.co/ken11/albert-base-japanese-v1}{ken11/albert-base-japanese-v1}& Ja & 11M & Rep. \\
        \href{https://huggingface.co/izumi-lab/bert-small-japanese-fin}{izumi-lab/bert-small-japanese-fin} \citep{Suzuki-etal-2023-ipm} & Ja & 18M & Rep.\\
        \href{https://huggingface.co/cl-tohoku/bert-base-japanese-whole-word-masking}{cl-tohoku/bert-base-japanese-whole-word-masking} & Ja & 111M & Rep.\\
        \href{https://huggingface.co/rinna/japanese-roberta-base}{rinna/japanese-roberta-base} \citep{rinna_pretrained2021} & Ja & 111M & Rep.\\
        \href{https://huggingface.co/cl-tohoku/bert-base-japanese-v2}{cl-tohoku/bert-base-japanese-v2} & Ja & 111M & Rep.\\
        \href{https://huggingface.co/xlm-roberta-base}{xlm-roberta-base} \citep{xlm-roberta} & Ja & 278M & Rep.\\
        \href{https://huggingface.co/Twitter/twhin-bert-base}{Twitter/twhin-bert-base} \citep{zhang2022twhin} & Ja & 279M & Rep.\\
        \href{https://huggingface.co/cl-tohoku/bert-large-japanese}{cl-tohoku/bert-large-japanese} & Ja & 337M & Rep.\\
        \href{https://huggingface.co/xlm-roberta-large}{xlm-roberta-large} \citep{xlm-roberta} & Ja & 560M & Rep.\\
        \href{https://huggingface.co/Twitter/twhin-bert-large}{Twitter/twhin-bert-large} \citep{zhang2022twhin} & Ja & 562M & Rep.\\
        \href{https://huggingface.co/monologg/koelectra-base-v3-generator}{monologg/koelectra-base-v3-generator} \citep{park2020koelectra} & Ko & 37M & Rep.\\
        \href{https://huggingface.co/klue/roberta-small}{klue/roberta-small} \citep{park2021klue} & Ko & 68M & Rep.\\
        \href{https://huggingface.co/snunlp/KR-FinBert}{snunlp/KR-FinBert} \citep{kr-FinBert} & Ko & 101M & Rep.\\
        \href{https://huggingface.co/beomi/kcbert-base}{beomi/kcbert-base} \citep{lee2020kcbert} & Ko & 109M & Rep.\\
        \href{https://huggingface.co/klue/bert-base}{klue/bert-base} \citep{park2021klue} & Ko & 111M & Rep.\\
        \href{https://huggingface.co/klue/roberta-base}{klue/roberta-base} \citep{park2021klue} & Ko & 111M & Rep.\\
        \href{https://huggingface.co/monologg/kobigbird-bert-base}{monologg/kobigbird-bert-base} & Ko & 114M & Rep.\\
        \href{https://huggingface.co/kykim/bert-kor-base}{kykim/bert-kor-base} \citep{kim2020lmkor} & Ko & 118M & Rep.\\
        \href{https://huggingface.co/lassl/bert-ko-base}{lassl/bert-ko-base} & Ko & 125M & Rep.\\
        \href{https://huggingface.co/klue/roberta-large}{klue/roberta-large} \citep{park2021klue} & Ko & 337M & Rep.\\
        \begin{tabular}{@{}c@{}}\href{https://huggingface.co/sentence-transformers/paraphrase-multilingual-mpnet-base-v2}{sentence-transformers/paraphrase-multilingual-mpnet-base-v2}  \\ \citep{sentencebert} \end{tabular}  & Ja & 278M & Allo. \\
        \begin{tabular}{@{}c@{}}\href{https://huggingface.co/ptaszynski/yacis-electra-small-japanese-cyberbullying}{ptaszynski/yacis-electra-small-japanese-cyberbullying} \\ \citep{shibata2022yacis-electra} \end{tabular} & Ja & 14M & Allo. \\
        \hline
\end{tabular}
\caption{Models used in this study. Shown are the language the model was used for (Lang.), 
the parameter count (Params.) and the application (App.) for which the model was used for.
Models were either used for studying representational (Rep.) or allocational (Allo.) biases.
}
\label{tbl:apx_LMs}
\end{table*}

\end{onecolumn}

\end{document}